\documentclass{article}

\usepackage{arxiv}
\usepackage{arabtex}
\usepackage[T1]{fontenc}    
\usepackage{hyperref}       
\usepackage{url}            
\usepackage{booktabs}       
\usepackage{amsfonts}       
\usepackage{nicefrac}       
\usepackage{microtype}      
\usepackage{lipsum}		

\usepackage{amssymb}

\usepackage{graphicx}
\usepackage[shortlabels]{enumitem}

\usepackage{verbatim}
\usepackage{float}
\usepackage{newfloat}

\renewenvironment{abstract}%
{\centerline{\large\bf Abstract}%
	\begin{list}{}%
		{\setlength{\rightmargin}{0.6cm}%
			\setlength{\leftmargin}{0.6cm}}%
		\item[]\ignorespaces}%
	{\unskip\end{list}}

\title{\emph{ANETAC}: Arabic Named Entity Transliteration and Classification Dataset}

\author{
  Mohamed Seghir Hadj Ameur\thanks{Corresponding author. Feel free to contact me via my personal email mohamedhadjameur@gmail.com} \\
  Department of Computer Science\\
  USTHB University\\
  Bab-Ezzouar, Algiers, Algeria\\
  \texttt{mhadjameur@usthb.dz} \\
   \And
  Farid Meziane\\
  Informatics Research Centre\\ 
  University of Salford\\ 
  M5 4WT, United Kingdom\\
  \texttt{f.meziane@salford.ac.uk} \\
  \And
    Ahmed Guessoum\\
	Department of Computer Science\\
	USTHB University\\
	Bab-Ezzouar, Algiers, Algeria\\
	\texttt{aguessoum@usthb.dz} \\
}

\let\oldmaketitle\maketitle
\renewcommand{\maketitle}{\oldmaketitle\setcounter{footnote}{0}}

\usepackage{utf8}
\begin{document}
\setcode{utf8}
\maketitle

\begin{abstract}
In this paper, we make freely accessible \textit{ANETAC}\footnote{The ANETAC dataset is freely available on Github \url{https://github.com/MohamedHadjAmeur/ANETAC}.} our English-Arabic named entity transliteration and classification dataset that we built from freely available parallel translation corpora. The dataset contains $79,924$ instances, each instance is a triplet $(e, a, c)$, where $e$ is the English named entity, $a$ is its Arabic transliteration and $c$ is its class that can be either a Person, a Location, or an Organization. The \textit{ANETAC} dataset is mainly aimed for the researchers that are working on Arabic named entity transliteration, but it can also be used for named entity classification purposes. This dataset was developed and used as part of a previous research study done by Hadj Ameur et al. \cite{ameur2017arabic}. 
\end{abstract}


\section{Introduction}
The task of transliteration is the process of converting words (e.g. named entities) that are written in one language alphabet to another language that has a different alphabet while still preserving the phonetics of the transliterated words.  
One of the main difficulties when attempting to transliterate named entities from a given source language to another is the lack of some phonetic character correspondences.  For example, in the task of named entity transliteration between Arabic and English, several Arabic letters such as \textquotedblleft \< ث >\textquotedblright and \textquotedblleft \< ظ >\textquotedblright $ $ do not have direct single-letter correspondences in the English language alphabet. Table \ref{fig:error_table2} presents some English named entities and their transliteration in the Arabic language.

\begin{table}[h]
	\centering
	\caption{English named entities and their equivalent Arabic transliterations}
	{\def\arraystretch{1.05}\tabcolsep=10pt	
	\begin{tabular}{c c}
		\hline
		\textbf{English} & \textbf{Arabic}  \\ 
		\hline
		
		Brandes &  \< برانديس >   (Brandees)\\
		Mayhawk & \< مايهوك >   (Mayhouk) \\ 
		Cressner & \< كريسنير >   (Crissneer)\\
		Husseini & \< حسيني >  (Husseini)\\
		\hline
	\end{tabular} 
	}
	\label{fig:error_table2}
\end{table}

Accurate transliteration of named entities is useful for several applications such as machine translation \cite{hermjakob2008name,habash2008four}, and cross-lingual information retrieval \cite{virga2003transliteration,fujii2001japanese}. Though a great deal of attention has been devoted to improving this task for many languages such as English, only limited studies have been made with regard to Arabic mainly due to the lack of transliteration datasets. In this paper, we make accessible \textit{ANETAC}, an English-Arabic named entity transliteration and classification dataset that we built from freely available parallel translation corpora. It contains 79,924 English-Arabic named entities along with their respective classes that can be either a Person, a Location, or an Organization. Table \ref{fig:stat} shows statistics about the \textit{ANETAC} named entities classes.

\begin{table}[h]
	\caption{Statistics about the number of named entities belonging to each class \cite{ameur2017arabic}}
	{\def\arraystretch{1.05}\tabcolsep=10pt
	\begin{tabular}{c c}
		\hline 
		\textbf{Named entity} & \textbf{Count} \\ 
		\hline 
		Person & 61,662 \\ 
		
		Location & 12,679 \\ 
		
		Organization & 5,583 \\
		
		\textbf{All} & 79,924 \\
		\hline 
	\end{tabular} 
	}
	\centering
	\label{fig:stat}
\end{table}

To make it easier for other researchers to train and compare their own models, the \textit{ANETAC} dataset is divided into training, development, and test sets as shown in Table \ref{fig:sets}.

\begin{table}[h]
	\caption{Instance counts in the train, development and test datasets of our transliteration corpus \cite{ameur2017arabic}}
	{\def\arraystretch{1.05}\tabcolsep=10pt
	\begin{tabular}{c c c c}
		\hline 
		\textbf{Sets} & \textbf{Train}  & \textbf{Dev}  & \textbf{Test} \\ 
		\hline 
		Named entities count & 75,898 &  1004 & 3013 \\ 
		\hline 
	\end{tabular} 
	}
	\centering
	
	\label{fig:sets}
\end{table}

As pointed out by many recent studies \cite{ameur2017arabic,rosca2016sequence}, there is a lack of Arabic machine transliteration datasets. To the best of our knowledge, there is only one freely available English-Arabic transliteration dataset that contains no more than 12,877 pairs \footnote{\url{https://github.com/google/transliteration}}, thus, we believe that our dataset will be a valuable addition. The importance of the \textit{ANETAC} dataset can be summarized as follows:
\begin{itemize}
	\item	This dataset is useful for many applications such as (1) training state-of-the-art English-Arabic machine transliteration models, (2) training Arabic named entity classification models, (3) handling Out-Of-Vocabulary (OOV) words in machine translation, (4) dealing with proper names in Cross-lingual Information Retrieval.
	\item	This dataset is mainly aimed for those researchers working on Arabic named entity transliteration, but it can also be used for named entity classification purposes.
	\item	This dataset also contains a test set that can be used as a benchmark to compare the results of English-Arabic transliteration systems. First transliteration results have been already reported on this test set by Hadj Ameur et al. \cite{ameur2017arabic} and will be shown in Section \ref{Baseline}.
\end{itemize}

In the remainder of this paper, section \ref{Dataconst} presents the corpus construction methodology that we adopted in the development of this dataset. Section \ref{Baseline} presents the baseline transliteration results that have been obtained using the \textit{ANETAC} dataset.  Finally, section \ref{conclusion} provides a conclusion to this paper.

\section{Building a Transliteration Corpus}
\label{Dataconst}

\begin{figure}[h]
	\centering
	\includegraphics[width=0.5\linewidth]{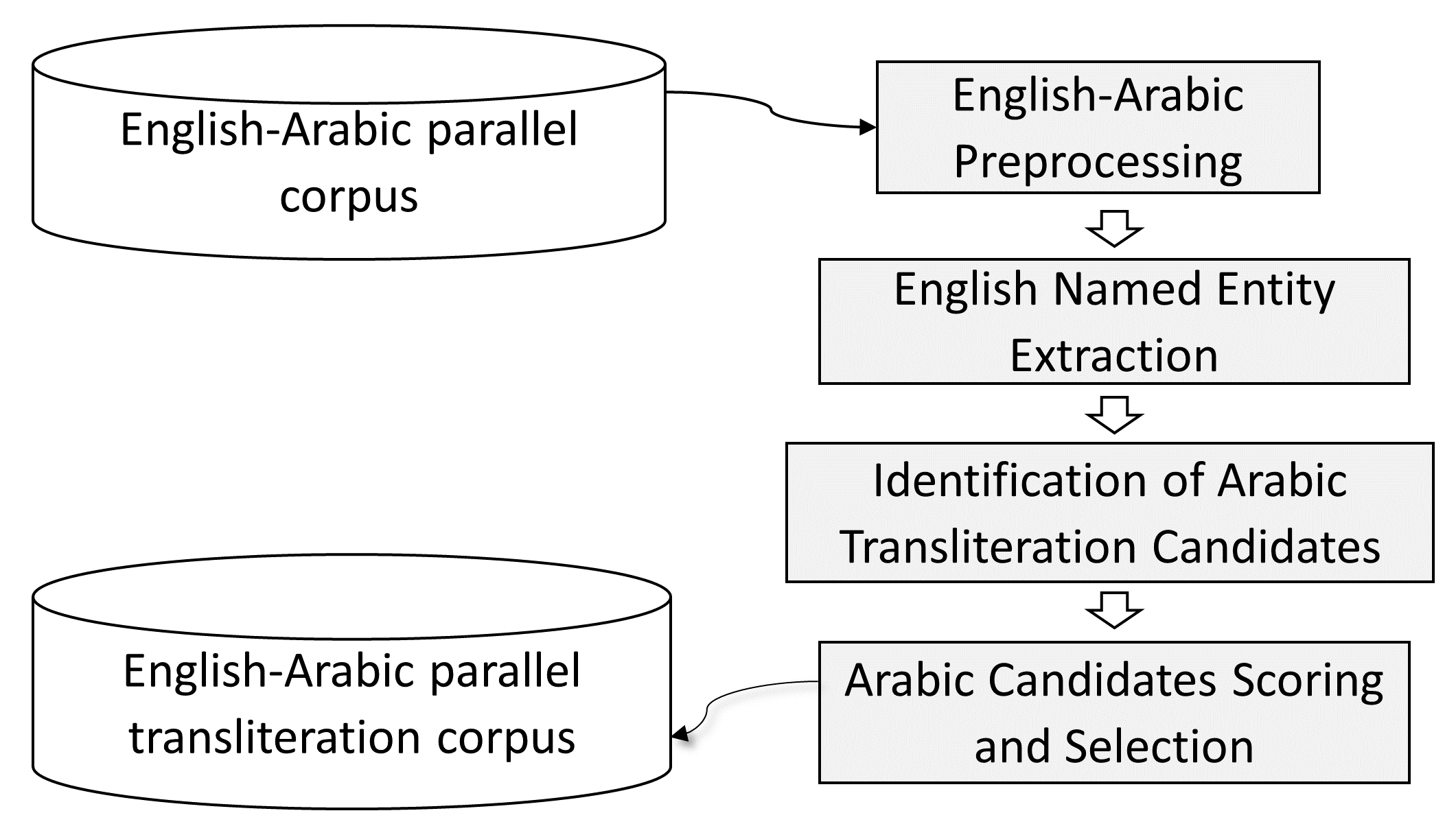}
	\caption{Architecture of our parallel English-Arabic Named entity extraction system \cite{ameur2017arabic}}
	\label{fig:extraction_system}
\end{figure}

As stated in the original work of Hadj Ameur et al. \cite{ameur2017arabic}\footnote{We note that this description of the extraction system is mostly based on the original paper of Hadj Ameur et al. \cite{ameur2017arabic}.}, the extraction system (see Fig. \ref{fig:extraction_system}) uses freely available parallel corpora\footnote{The English-Arabic parallel corpora that we used are available on the opus website: \url{http://opus.nlpl.eu}.}  in order to automatically extract bilingual named entities. The English-Arabic corpora that we have used are provided in Table \ref{fig:original_corpus_statistics}.

\begin{table}[h]
	\centering
	\caption{Statistics about the used English-Arabic parallel corpora \cite{ameur2017arabic}}
	{\def\arraystretch{1.05}\tabcolsep=10pt
		\begin{tabular}{c c}
			\hline
			\textbf{Corpus} & \textbf{Sentences (in millions)} \\ 
			\hline
			United Nation  & 10.6M \\ 
			Open Subtitles  & 24.4M \\ 
			News Commentary  & 0.2M \\ 
			IWSLT2016 & 0.2M \\
			\textbf{All} & 35.4M \\
			\hline
		\end{tabular} 
	}
	\label{fig:original_corpus_statistics}
\end{table}

As shown in Fig. \ref{fig:extraction_system}, the system starts by a preprocessing phase in which the English and Arabic sentences are tokenized and normalized. Then, the English named entities are identified in each sentence belonging to the English-side of the parallel corpus. A set of Arabic transliteration candidates will then be associated with each English named entity. Finally, the best Arabic transliteration candidate will be selected for each English named entity. The detail of these step are provided in the remainder of this section. 

\subsection{Parallel Named Entity Extraction}
The ultimate goal is to extract the correct Arabic transliteration of each English named entity. Given a corpus of English-Arabic parallel sentences $S=\{(e_1,a_1),...,(e_m,a_m)\}$, we use the Stanford English Named Entity Recognizer \cite{finkel2005incorporating} to find all the English named entities that are present in the parallel corpus $E_{ne}=\{n_1,n_2,...,n_k\}$, where $k$ is the total number of named entities. Since each singleton word belonging to a multi-word English named entity can always be transliterated solely without needing its context, we decomposed all the English named entities containing multiple words to several singleton entities.  For each English named entity $n_i$ belonging to an English sentence $e_j$, we end up with a list of pairs $(n_i,a_j)$ denoting that the $i^{th}$ English named entity (singleton word) is associated with the $j^{th}$ Arabic sentence.  

\subsection{Candidates Extraction and Scoring}
The previous step leaves us with a set of pairs  $(n_i,a_j)$, where $n_i$ is the English named entity (word) and $a_j$ is the Arabic sentence containing its transliteration. To find the correct transliterated word of $n_i$ in the Arabic sentence $a_j$, we first removed all the frequent Arabic words from it using a vocabulary containing the top $n$ most frequent Arabic words, with $n = 40000$, that we built automatically from our parallel corpus. This ensures that the remaining words in the Arabic sentence $a_j$ are mostly rare words. All the remaining words in $a_j$ are considered as transliteration candidates $C(a_j)=\{c_{j1},c_{j2},...,c_{jt}\}$, where $c_{ji}$ denotes the $i^{th}$ candidate word found in the $j^{th}$ Arabic sentence, and $t$ is the total number of Arabic candidates in $C(a_j)$. We used the transliteration tool available in the polyglot multilingual NLP library\footnote{\url{https://github.com/aboSamoor/polyglot}} to obtain an approximate Arabic transliteration $t_i$ of each English named entity $n_i$. For each English named entity $n_i$ having the approximate transliteration $t_i$ and the list of Arabic candidates $C(a_j)$, the score of each Arabic candidate is estimated using the following three features:

\begin{enumerate}
	\item The total number of shared characters: this feature takes into account the count of shared characters between each Arabic candidate in $C(a_j)$ and the approximate transliteration $t_i$.
	
	\item  The longest shared sequence: this feature takes into account the length of the longest common sequence of characters between each Arabic candidate in $C(a_j)$ and the approximate transliteration $t_i$.
	
	\item  Length difference penalty: this feature is used to penalize the  $C(a_j)$ candidates according to their level of dissimilarity with the approximate transliteration $t_i$.
\end{enumerate}

The final score of each candidate is then estimated by averaging the score of all the three features. The candidate having the highest score is then selected if its corresponding final score surpasses a certain confidence threshold. Some examples of the extracted English-Arabic named entities are provided in Table \ref{fig:exemple_entities_tz}. The reader should recall that the Arabic language has no letters for the English sound \textquotedblleft v\textquotedblright $ $, \textquotedblleft p\textquotedblright $ $ and \textquotedblleft g\textquotedblright.

	\begin{table}[h]
	\centering
	\caption{Some examples of the extracted English-Arabic named entities \cite{ameur2017arabic}}
	{\def\arraystretch{1.1}\tabcolsep=10pt
	\begin{tabular}{c c c}
		\hline
		\textbf{Entity class} & \textbf{English} & \textbf{Arabic} \\ 
		\hline
		
		PERSON & Villalon  &	\< فيلالون > (filaloun)\\ 
		LOCATION &  Nampa & \< نامبا > (namba)\\
		ORGANIZATION &  Soogrim & 	\< سوغـريم > (soughrim)\\
		\hline
	\end{tabular} 
	}
	\label{fig:exemple_entities_tz}
\end{table}

\section{Baseline Results}
\label{Baseline}
This section provides the English-to-Arabic and Arabic-to-English baselines' transliteration results that we have obtained when using the  \textit{ANETAC} dataset for both the training and testing of our models \cite{ameur2017arabic}.  The baseline results (Table \ref{fig:test-en-ar}) are reported in terms of both Word Error Rate (WER) and Character Error Rate (CER) on the \textit{ANETAC} test set\footnote{\url{https://github.com/MohamedHadjAmeur/ANETAC}}.  

\begin{table}[h]
	\centering
	\caption{Baseline transliteration results in terms of WER and CER reported on the \textit{ANETAC} test set}
	{\def\arraystretch{1.1}\tabcolsep=10pt
		\begin{tabular}{c c c}
			\hline 
			\textbf{Directions} & \textbf{WER}  & \textbf{CER}   \\ 
			\hline 
			English-to-Arabic & \textbf{5.40 }&  \textbf{0.95}  \\
			Arabic-to-English & \textbf{65,16 }&  \textbf{16.35}  \\
			\hline 
		\end{tabular} 
	}
	\label{fig:test-en-ar}
\end{table}	

As shown in Table \ref{fig:test-en-ar}, the results of the Arabic-to-English transliteration are still poor, thus much work is still needed to improve them. We note the baseline models that we have used are based on the attention-based encoder-decoder architecture \cite{bahdanau2014neural} and trained at the character level. 

\section{Conclusion}
\label{conclusion}
In this work, we have made accessible the \textit{ANETAC} dataset, that we developed as part of our previous work \cite{ameur2017arabic}. We have shown how this dataset is built from parallel translation corpora by relying on several features and tools. We also presented the baseline results that we have achieved on the tasks of English-to-Arabic and Arabic-to-English machine transliteration.  We encourage all researchers that are interested in this task to try and achieve better results. Finally, we hope that this dataset will have a positive impact on the current state of Arabic-English named entity transliteration.



\end{document}